\documentclass[letterpaper, 10 pt, conference]{ieeeconf}  

\IEEEoverridecommandlockouts                              

\usepackage{graphics} 
\usepackage{epsfig} 
\usepackage{mathptmx} 
\usepackage{times} 
\usepackage{amsmath,algorithm} 
\usepackage{amssymb}  
\usepackage{mathtools}
\usepackage{stfloats}
\usepackage[noend]{algpseudocode}
\usepackage{graphicx}
\usepackage{multirow}
\usepackage{multicol}
\usepackage{caption}
\usepackage{subcaption}
\usepackage{url}
\usepackage[bookmarks=false]{hyperref}
\usepackage{censor}

\hypersetup{
colorlinks=true,
linkcolor=blue,
filecolor=magenta,
urlcolor=cyan,
}

\DeclareMathOperator{\sign}{sign}

\graphicspath{{figures/}}

\title{\LARGE \bf
Cooperative Transport of Unknown Objects by Distributed Effort Regulation in Multi-agent Systems
}

\author{Tamzidul Mina$^{1*}$, Shyam S. Kannan$^{2*}$, Wonse Jo$^{3*}$, Shaocheng Luo$^{4*}$, Galen B. King $^{5}$, and Byung-Cheol Min$^{6*}$
\thanks{$^{1}$ Sandia National Laboratories, Albuquerque, NM 87123.
        {\tt\small tmina@sandia.gov}}%
\thanks{$^{2}$ The Boeing Company, Bengaluru, India. 
        {\tt\small shyamsundar.kannan@boeing.com}}%
\thanks{$^{3}$ Department of Information and Telecommunication, Incheon National University, South Korea.
        {\tt\small jow@inu.ac.kr}}%
\thanks{$^{4}$ Dept. of Electrical and Computer Engineering, Duke University.
        {\tt\small shaocheng.luo@duke.edu}}%
\thanks{$^{5}$ School of Mechanical Engineering, Purdue University, West Lafayette, IN 47907, USA 
{\tt\small kinggb@purdue.edu}}
\thanks{$^{6}$ Department of Computer Science and Department of Intelligent Systems Engineering, Indiana University, Bloomington, IN, USA.
        {\tt\small minb@iu.edu}}%
\thanks{$^{*}$ This research in its entirety was conducted while previously affiliated with the SMART Lab, Department of Computer and Information Technology, Purdue University, and the School of Mechanical Engineering, Purdue University, West Lafayette, IN 47907, USA. }
}

\begin{document}

\maketitle
\thispagestyle{empty}
\pagestyle{empty}

\begin{abstract}

A distributed multi-robot object transportation strategy is proposed for objects of unknown shape, size, geometry, mass and mass distribution without any prior knowledge about the object or number of robots involved in the transportation process. 
Robots follow the boundary of the object and dynamically determine viable pushing points. 
A novel object-model-free heading-error-driven pushing effort regulation method is proposed to promote goal-directed object motion; robots regulate their pushing effort using individual and collective heading errors.
Robots dynamically relocate around the object boundary if a pushing point is lost or the required pushing effort exceeds its capability to correct its heading. 
The proposed method is validated using simple and complex shapes with convex and non-convex boundary regions and dynamically changing weight distribution of objects. 
Detailed analysis on the effects of robot distribution and initial object pose on the transportation performance in terms of required time and path efficiency is also presented. 
Scalability analysis revealed that object transportation time is dependent on the number of pushing surfaces available which in turn depends on the object size, shape, and set inter-robot safety distance. 
The proposed system yielded highly efficient object paths with consistent elapsed transport times over several trials robust to external disturbances such as non-uniform friction between contacting surfaces and robot wheel slip. 

\end{abstract}

\section{Introduction}\label{sec:intro}
Object transportation by robotic systems has a rich potential for application in warehouse logistics, path clearing in search and rescue, evacuation scenarios, and the construction industry. 
Multi-robot teams can provide redundancy in enabling reconfiguration, distributed sensing and action at scales that are impractical or expensive with a single unit.
Traditional approaches often frame the necessity of multi-robot configurations around the pooling of physical force to move large payloads.
In contrast, this work emphasizes that the primary value of a multi-robot system lies in utilizing multiple independent contact locations to enable the distributed regulation of object motion despite unknown geometry, unknown center of mass, non-uniform surface friction, and continuous external disturbances.

Multi-robot object transportation can generally be categorized into caging/grasping, towing, and pushing strategies. 
Caging and grasping require a minimum number of robots, graspable features, and are limited to convex shapes \cite{tejada2025enhancing, wan2017multirobot}, while towing requires attachable points for cables \cite{du2023distributed}. 
Pushing strategies offer versatility and robustness \cite{alkilabi2017cooperative} but face inherent goal-occlusion challenges. 
While occlusion-leveraging strategies work for simple geometries \cite{chen2015occlusion}, complex shapes often cause significant drift making identification of correct pushing surfaces critical for path efficiency.

\begin{figure}[t]
	\centering
	\includegraphics[width=0.97\linewidth]{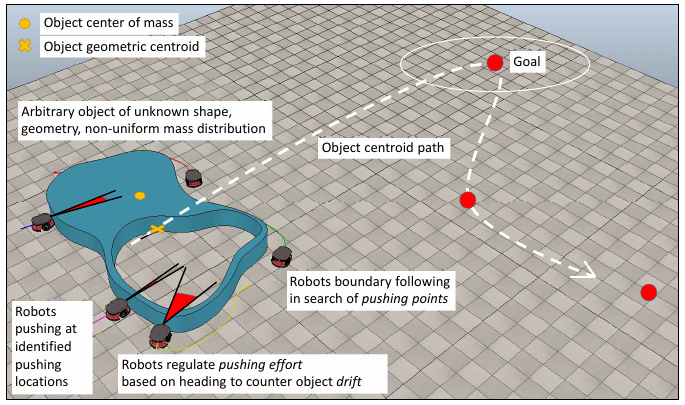}
    \vspace{-2mm}
  	\caption{Conceptual illustration of the proposed multi-robot arbitrary object transportation approach.}
    \vspace{-6mm}
  	\label{fig:intro}
\end{figure}

Pushing strategies often integrate force feedback for motion contribution and path keeping \cite{neumann2016hybrid}.
Coordination ranges from centralized systems that lack fault tolerance \cite{neumann2014object} to decentralized leader-follower models requiring no communication \cite{wang2016kinematic}. 
Analytical pushing strategies \cite{lee2015hierarchical} and data-driven models \cite{inceoglu2018continuous} have been explored, yet they remain limited by rigid assumptions \cite{stuber2020let}. 
Recent advancements utilizing Distributed Model Predictive Control (DMPC) \cite{muhammed2026real}, multi-objective optimization under distributed control \cite{tian2025multi}, rigid formation architectures \cite{bai2025multi}, or event-driven collaborative frameworks \cite{felix2026event} introduce heavy computational overhead, depend heavily on strict prior object geometry tracking, or require explicit modeling constraints to safely capture payload dynamics. Consequently, combining applied force regulation requirements with fully distributed, model-free coordination strategies that adapt to arbitrary object physical configurations online has been a significant challenge \cite{rosenfelder2024force}.

In this paper, we propose a scalable multi-robot pushing strategy for transporting objects of unknown shape, size, mass, and center of mass. 
The method requires no prior knowledge of the object or robot count and remains robust against imprecise contact and non-uniform friction effects. 
Using individual assessments within a limited field-of-view (FOV), robots:
\begin{itemize}
    \item identify potential pushing surfaces on the perceived object boundary and initiate pushing minimizing the risk of physical damage to the object, and
    \item regulate their \textit{pushing effort} based on a proposed object-model-free strategy such that their combined action creates a resultant motion of the unknown object in the intended direction.
\end{itemize}
Robots are assumed to share heading error information through lightweight communication; however, pushing surface identification and effort regulation are computed individually, preserving decentralized control.
 Cooperation in our framework is realized through distributed effort adaptation: each robot independently adjusts its pushing effort using the collective heading error shared by currently engaged robots.
This object-model-free regulation coordinates the robots' pushing efforts to promote goal-directed motion despite mass distribution variations or external disturbances.

\section{Transport strategy}\label{sec:method}
A bounded planar environment contains an arbitrary shaped object $O$ of unknown mass and mass distribution to be transported to goal $g \in \mathbb{R}^2$. 
Each robot $R_i, i \in\{1,\ldots,N\}$ with position and orientation/heading defined as $r_i \in \mathbb{R}^2$ and $\pi\geq \theta_i >-\pi$ in the transport process, is assumed to have a circular FOV with detection radius $d_r$ and is aware of the direction towards $g$ defined as $\theta_{i,g}$ relative to the robot local frame.
\textit{Pushing effort} of a robot is defined as a unit-less measure of applied pushing force relative to the pushing ability of the robot platform; it is regulated based on individual and the collective heading error of all robots during pushing to ensure the object is pushed in the intended direction. 
Robots relocate to a new pushing location if the required pushing effort exceeds its own ability. 
Details of the proposed object-model-free pushing effort regulation method are provided in Section \ref{sec:push_effort}.

\subsection{Finite State Machine Controller}\label{sec:fsm}
\begin{figure}[t]
	\centering
	\includegraphics[width=0.80\linewidth]{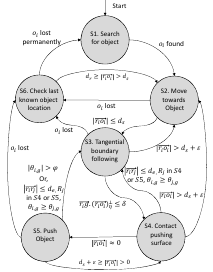}
    \vspace{-2mm}
  	\caption{Finite state machine controller with initial state $S1$.}
    \vspace{-5mm}
  	\label{fig:fsm}
\end{figure}

\begin{figure*}[t]
\centering
    \includegraphics[width=0.95\linewidth, height=2.65in]{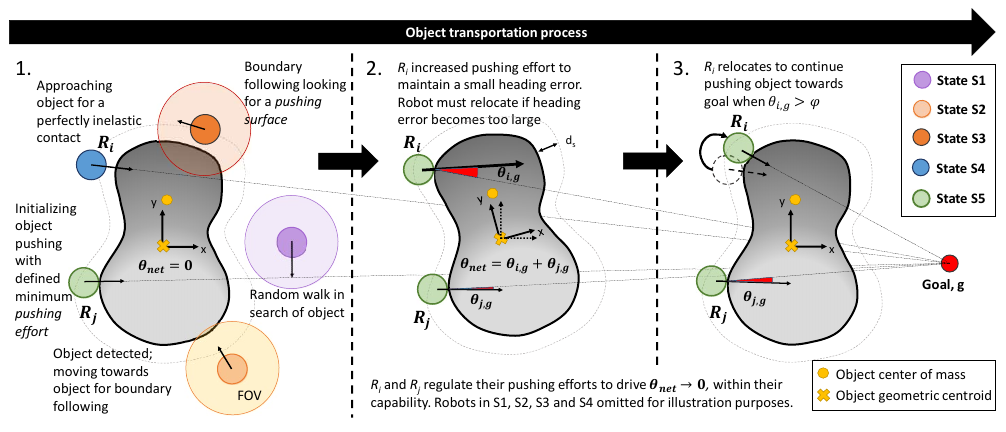}
\vspace{-2mm}
  	\caption{Conceptual image of the proposed multi-robot object transportation process for objects with unknown geometric centroid and center of mass (heavy and light ends illustrated by darker and lighter shade). 
  	}
    \vspace{-3mm}
\label{fig:concept}
\end{figure*}

The object transportation process is implemented as a finite state machine as shown in Fig. \ref{fig:fsm} with a conceptual illustration shown in Fig. \ref{fig:concept}. 
At initial time, all robots search for the object with random walk at state $S1$. 
\begin{itemize}
    \item 
    \textbf{State \emph{S}1: Search for object.} $R_i$ in $S1$ performs random walk in the environment avoiding collision and looking for the object $O$. Grid based search algorithms \cite{ryu2011search,mannadiar2010optimal} may be implemented for $S1$; random walk is chosen here for simplicity. The set of detected boundary points of object $O$ within $d_r$ of $R_i$ is recorded as ordered pairs of distance and relative direction from $R_i$; the nearest boundary point is identified as $\{o_{i},\theta_{i,o}\}$, where $o_i$ is the minimum distance and $\theta_{i,o}$ is the relative direction of $o_i$ from $R_i$. Transition:
    \begin{itemize}
        \item $S1 \longrightarrow S2$: if $o_{i}$ found.
    \end{itemize}
    \item \textbf{State \emph{S}2: Move towards object \emph{O}.} $R_i$ in $S2$ translates towards $o_{i}$ to reach a defined boundary following distance $d_s$. Transitions:
    \begin{itemize}
        \item $S2 \longrightarrow S3$: if $|\overrightarrow{r_io_i}|\leq d_s$ for $R_i$ in $S2$.
        \item $S2 \longrightarrow S6$: if $o_{i}$ lost from FOV.
    \end{itemize}
    \item \textbf{State \emph{S}3: Follow object boundary until a pushing surface is detected.} Robot $R_i$ in $S3$ determines the collective heading angle error of all robots in $S5$ as, $\theta_{net}=\sum_{i=1, R_i \text{ in } S5}^{N} \theta_{i,g}$, where $\theta_{i,g}$ is the heading angle error of $R_i$ when pushing in $S5$, to set $\eta=\sign(\theta_{net})$. $\eta$ determines the direction of tangential boundary motion at a distance $d_s$ from the nearest detected point on the object $o_{i}$ defined as $(\widehat{\overrightarrow{r_io_{i}}})^{\perp}_\eta$. During boundary following, $R_i$ assesses $\widehat{\overrightarrow{r_ig}} \cdot (\widehat{\overrightarrow{r_io_{i}}})^{\perp}_\eta$ looking for a pushing surface point, where $\widehat{\overrightarrow{r_ig}}$ is the unit vector from $R_i$ to the goal $g$. Transitions:
    \begin{itemize}
        \item $S3 \longrightarrow S2$: if $|\overrightarrow{r_io_{i}}|>d_s+\epsilon$ where $\epsilon$ is a defined small tolerance distance.
        \item $S3 \longrightarrow S4$: if $\widehat{\overrightarrow{r_ig}} \cdot (\widehat{\overrightarrow{r_io_{i}}})^{\perp}_\eta \leq \delta$ detected for $\delta \approx 0$ with $g$ obstructed from $R_i$ by $O$, a viable pushing surface point is reached.
        \item $S3 \longrightarrow S6$: if $o_{i}$ lost from FOV.
    \end{itemize}
    \item \textbf{State \emph{S}4: Contact pushing surface at $\pmb{o_i}$.} Robot $R_i$ in $S4$ rotates towards the object and approaches the object surface at the determined pushing point $o_{i}$ and makes a perfectly inelastic collision with the object; i.e. collision with a coefficient of restitution, $e=0$. Transition:
    \begin{itemize}
        \item $S4 \longrightarrow S2$: if $|\overrightarrow{r_io_{i}}|>d_s+\epsilon$.
        \item $S4 \longrightarrow S3$: if $|\overrightarrow{r_ir_j}|\leq d_e$, $R_j$ in $S4$ or $S5$ with $|\theta_{i,g}| \geq |\theta_{j,g}|$, where $d_e$ is a set inter-robot safety distance.
        \item $S4 \longrightarrow S5$: if $|\overrightarrow{r_io_{i}}|\approx 0$.
    \end{itemize}
    \item \textbf{State \emph{S}5: Push object at $\pmb{o_i}$ regulating effort.} Robot $R_i$ in $S5$ pushes the object at the determined pushing point $o_i$ with a pushing effort magnitude defined as $\rho_5=f(\theta_{net},\theta_{i,g})$, (see Section \ref{sec:push_effort}) in the direction of the object goal position $g$, while $\theta_{i,g}$ is bounded within $[-\phi,\phi]$. Transition:
    \begin{itemize}
        \item $S5 \longrightarrow S3$: if $|\theta_{i,g}|>\phi$ or $|\overrightarrow{r_ir_j}|\leq d_e$ $R_j$ in $S4$ or $S5$ with $|\theta_{i,g}| \geq |\theta_{j,g}|$ or no net motion of object with $\theta_{net} \approx 0$ (possible stagnation point).
        \item $S5 \longrightarrow S4$: if $d_s+\epsilon > |\overrightarrow{r_io_{i}}|>0$.
        \item $S5 \longrightarrow S6$: if $o_i$ lost from FOV.
    \end{itemize}
    \item \textbf{State \emph{S}6: Check last known direction for object \emph{O}.} In each state $S2$, $S3$ and $S5$ robot $R_i$ records the last known direction of $o_i$ as $\theta_{i,o}^{prev}$. $R_i$ in $S6$ rotates towards the last known direction $\theta_{i,o}^{prev}$ and translates up to distance $d_r$ to reacquire the lost object $O$. Transition:
    \begin{itemize}
        \item $S6 \longrightarrow S1$: if $o_{i}$ lost permanently.
        \item $S6 \longrightarrow S2$: if $d_r \geq |\overrightarrow{r_io_{i}}|>d_s$.
    \end{itemize}
\end{itemize}

During boundary following in $S3$, if $R_i$ encounters a robot in $S5$ on its path, it treats the robot in $S5$ as part of the object and continues to follow around its boundary. However, the transition to $S4$ is temporarily disabled until $o_i$ is reacquired.

When object $O$ is far away from $g$, $\theta_{i,g}$ remains small with changes in $||r_i-g||$. Closer to $g$, $\theta_{i,g}$ changes rapidly and hence its defined bounds $[-\phi,\phi]$ determine how frequently robots will relocate. The constant $\phi$ must be chosen as small as possible such that $\theta_{net}$ remains close to zero after transient. Robot anti-wheel slip capabilities in achieving calculated pushing effort must also be considered when choosing $\phi$. 

\subsection{Motion Control}\label{sec:kinematics}
A potential field approach \cite{khatib1986real} is used to generate the movement vectors of robot $R_i$ at each defined state of the proposed finite state machine controller. The set of control inputs $u$ for each robot $R_i$ in each state is defined as:   
\vspace{-12mm}
\begin{multicols}{2}
  \begin{equation}
    u_{S1} = \rho_1 v_m \widehat{\overrightarrow{r_{i, rand}}}
  \end{equation}\break
  \begin{equation}
    u_{S2} = \rho_2 |\overrightarrow{r_io_{i}}|\widehat{\overrightarrow{r_io_{i}}}
  \end{equation}
\end{multicols}
\vspace{-4mm}
\begin{equation}
u_{S3} = \rho_{3} (|\overrightarrow{r_io_{i}}|)(\widehat{\overrightarrow{r_io_{i}}})^{\perp}_\eta + \frac{\rho_2}{2} (|\overrightarrow{r_io_{i}}|-d_s)\widehat{\overrightarrow{r_io_{i}}}
\end{equation}
\vspace{-16mm}
\begin{multicols}{2}
  \begin{equation}
    u_{S4} =\rho_{4} (|\overrightarrow{r_io_{i}}|)\widehat{\overrightarrow{r_io_{i}}}
  \end{equation}\break
  \begin{equation}
    u_{S5} =\rho_{5} v_{m}\widehat{\overrightarrow{r_{i}g}}
  \end{equation}
\end{multicols}
\noindent where $\widehat{\overrightarrow{r_{i, rand}}}$ is a random unit vector with direction defined within $\pm 45^o$ of $\theta_i$ updated at a fixed interval; $0<\rho_1\leq 1$, $\rho_2$, $\rho_{3}$ and $\rho_4$ are constants. The control input for $S6$ is defined as a constant attractive potential in the direction of $\theta_{i,o}^{prev}$ with gain constant $\rho_6$ for a travel distance of up to $d_r$. 

$u_{S2}$ and $u_{S4}$ apply attractive potentials driving $R_i$ toward $o_{i}$. 

$u_{S3}$ directs $R_i$ along a tangential path around the boundary in the direction $\eta$.
The first term governs tangential following, while the second term corrects the distance from the boundary to maintain preset distance $d_s$.
All robots are constrained by maximum linear and rotational velocity defined as $v_{m}$ and $\omega_m$ respectively.

\subsubsection{Pushing Effort Regulation}\label{sec:push_effort}
When a robot transitions from state $S4$ to $S5$, it is initially positioned in contact with the object with its heading aligned with the goal location. 
Without any prior knowledge of the object centroid or center of mass, it starts to push with a minimum defined effort in the hopes that other robots are present or will position themselves to balance the rotational moment of object $O$. 
Individuals constantly monitor their individual heading error from the goal location $\theta_{i,g}$ and the collective heading error of all robots in $S5$ defined as $\theta_{net}$; if both share the same sign, $R_i$ linearly increases its effort to drive $\theta_{net}$ toward zero.
The proposed \textit{pushing effort} model is defined as,
\begin{equation}\label{eq:rho5}
\rho_5 = \max(0,\sign(\theta_{net})\frac{2\alpha}{\pi}\theta_{i,g})+\frac{\alpha}{2}
\end{equation}
where $0 < \alpha \leq 1$ is a constant, and $-\phi \leq \theta_{i,g} \leq \phi$.
The proposed model increases a robot's pushing effort above $\alpha/2$ when its individual heading error has the same sign as $\theta_{net}$. This heading-dependent effort asymmetry is intended to reduce the collective heading error without an explicit object model.
\begin{figure}[t]
\centering
	\includegraphics[width=0.86\linewidth]{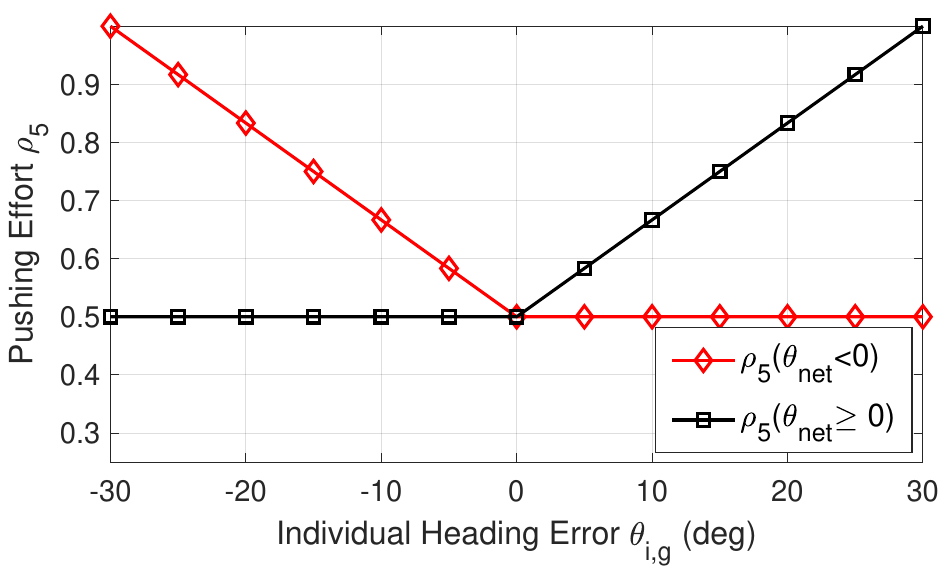}
    \vspace{-2mm}
  	\caption{Pushing effort $\rho_5$ model for robots in state $S5$ based on $\theta_{i,g}$ and $\sign(\theta_{net})$ for $\phi=30^o$, $\alpha=1$ following Eq. \ref{eq:rho5}.}
    \vspace{-5mm}
\label{fig:rho5}
\end{figure}
Acknowledging that in reality, individual robots are constrained by their pushing capabilities without any wheel slip and may be unable to drive the collective heading error to zero, this maximum pushing capability of each robot is reflected by the choice of $\alpha$. 
Robots default to half-capacity pushing unless adjusted depending on $\theta_{net}$, reserving the remainder for object directional corrections. 
$\alpha$ can also be adjusted dependent on wheel-ground friction conditions and required aggressiveness of the object transportation process.
When $|\theta_{i,g}|>\phi$, the robot transitions back to state $S3$ to search for another pushing location. This ensures that the collective heading error of all robots in $S5$ is always bounded within $[-n\phi,n\phi]$, where $n$ is the number of robots in $S5$ at any given time. An illustration of the proposed object-model-free pushing effort regulation for $\phi=30^o$ and $\alpha=1$ is shown in Fig. \ref{fig:rho5}. 

\subsubsection{Inter-robot Collision Avoidance}
Moving robots in the system repel one another to maintain a safe distance and prevent collision. Therefore, robot $R_i$ in $S1$, $S2$, $S3$ or $S6$ actively repels neighboring robot $R_j$ in $S1$, $S2$, $S3$, $S4$, $S5$ or $S6$ using a repulsive potential within safety distance $d_e$:
\begin{equation}\label{eq:repulsion}
u_{repel} = -\frac{\rho_r}{|\overrightarrow{r_ir_j}|^2}\widehat{\overrightarrow{r_ir_j}} \text{ for } |\overrightarrow{r_ir_j}|\leq d_e
\end{equation}
where $\rho_r$ is a repulsion constant.
$R_i$ in $S4$ or $S5$ prioritizes object pushing and relies on the repulsion set by other robots in its vicinity to maintain a safe distance; it transitions to $S3$ to avoid collision and relocates if a nearby pushing robot $R_j$ in $S4$ or $S5$ comes too close while pushing with a smaller heading angle error, i.e. $|\overrightarrow{r_ir_j}|\leq d_e$ for $R_i, R_j$ in $S4$ or $S5$ and $|\theta_{i,g}| \geq |\theta_{j,g}|$.

The proposed finite state machine and motion control methods allow robots to settle at identified pushing surfaces on the goal occluded region, push the object regulating their effort to correct individual heading error within the bounded angle $\phi$ while avoiding any inter-robot collisions. 
If no net object motion is observed while $\theta_{net}\approx0$, robots in $S5$ return to boundary following in $S3$ to search for alternative pushing locations.

\section{Validation}\label{sec:expt}
\begin{figure*}[t]
    \centering
    \begin{subfigure}[b]{1\textwidth}
      \includegraphics[width=\textwidth]{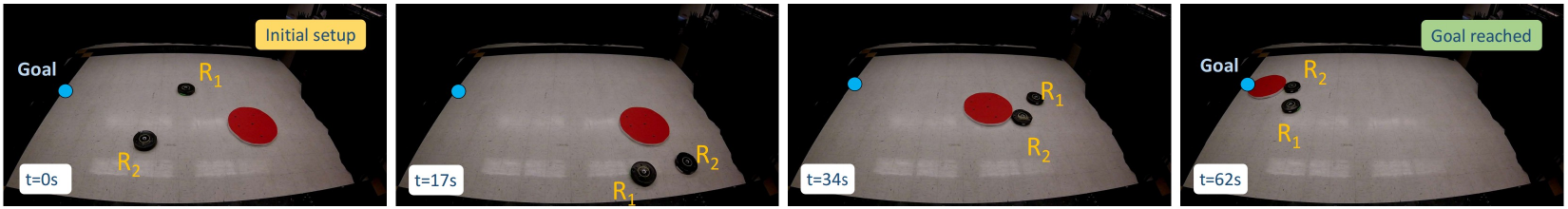}
      \vspace{-6mm}
        \caption{Trial $1$ time lapse illustration of the $2$ robots transporting a disk shaped object of uniform weight distribution towards the goal.}
        \label{fig:exptA1_tl}
        \vspace{3mm}
    \end{subfigure}
    \begin{subfigure}[b]{0.32\textwidth}
    \captionsetup{width=.95\linewidth}%
      \includegraphics[width=2.15in,height=1.0in]{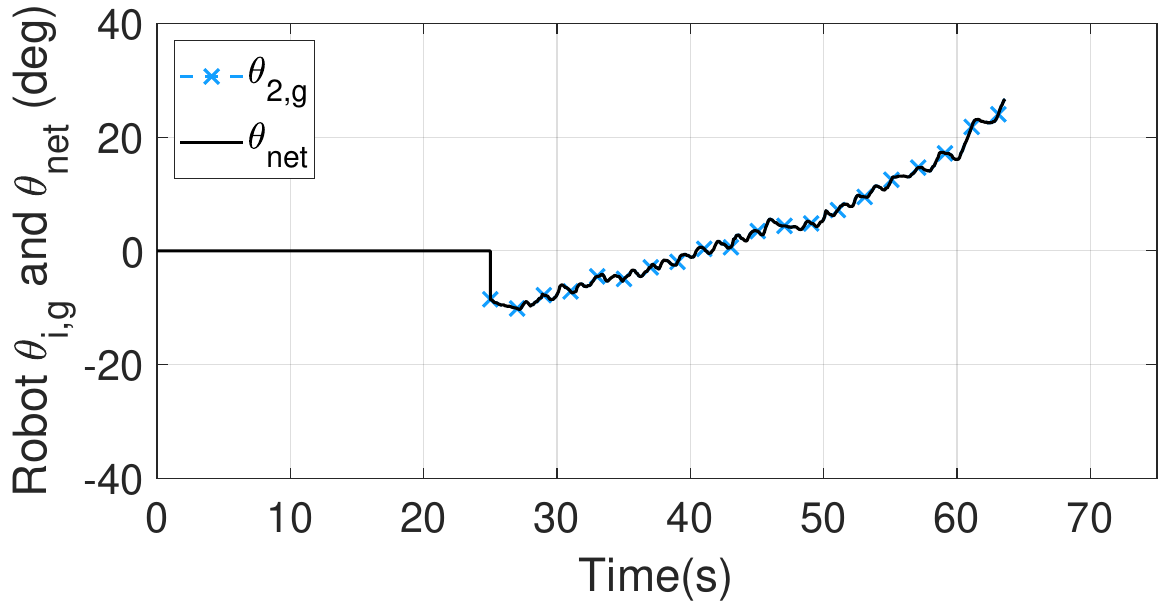}
        \caption{Individual robot and net heading angle error while pushing, confined within $[-\phi, \phi]$ and $[-2\phi, 2\phi]$ respectively.}
        \label{fig:exptA1_theta}
    \end{subfigure}
    \begin{subfigure}[b]{0.32\textwidth}
    \captionsetup{width=.95\linewidth}%
        \includegraphics[width=2.15in,height=1.0in]{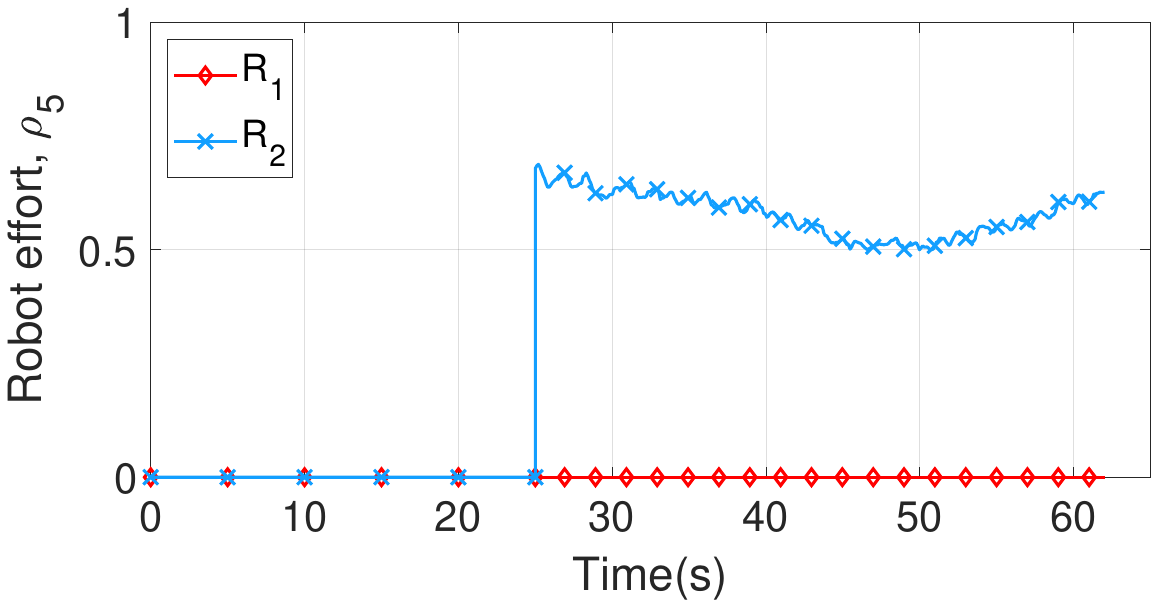}
        \caption{Pushing effort $\rho_5$ in response to individual and collective heading angle error to push object towards the goal location.}
        \label{fig:exptA1_effort}
    \end{subfigure}
    \begin{subfigure}[b]{0.32\textwidth}
    \captionsetup{width=.98\linewidth}%
        \includegraphics[width=2.15in,height=1.0in]{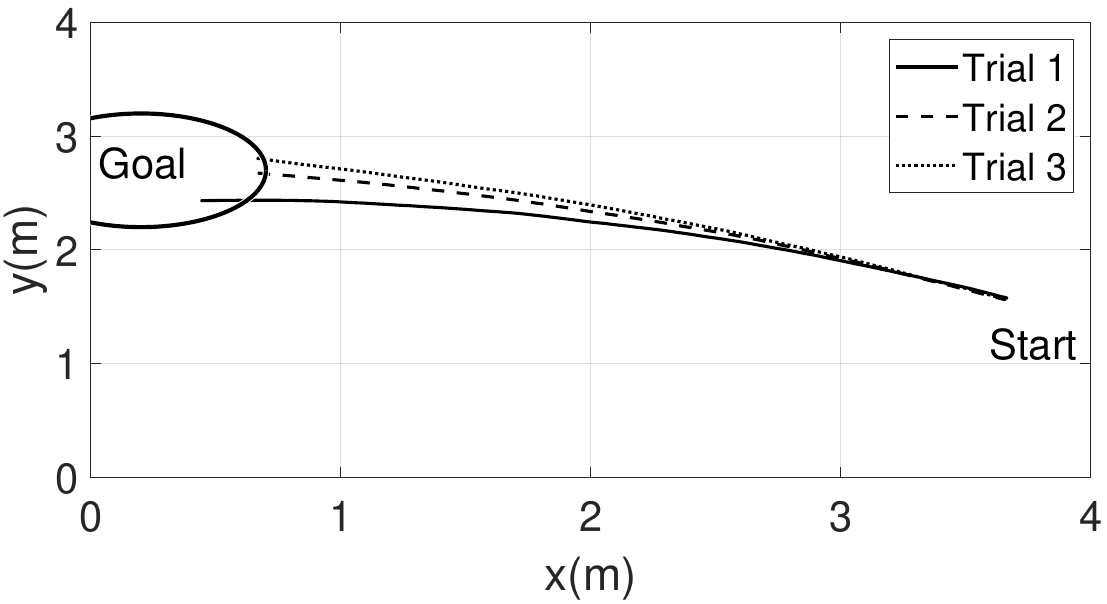}
        \caption{Object centroid path to goal, showing object path converging towards the goal region successfully for all trials.}
        \label{fig:exptA1_path}
    \end{subfigure}
    \caption{Experiment A.1: Transporting a $0.04$\,kg disk with diameter $0.83$\,m and uniform weight distribution using $2$ robots. With only one potential pushing location, the object was transported by one robot as expected.}
    \vspace{-4mm}
    \label{fig:exptA1}
\end{figure*}
\begin{figure*}[t]
\centering
    \begin{subfigure}[b]{1\textwidth}
      \includegraphics[width=\textwidth]{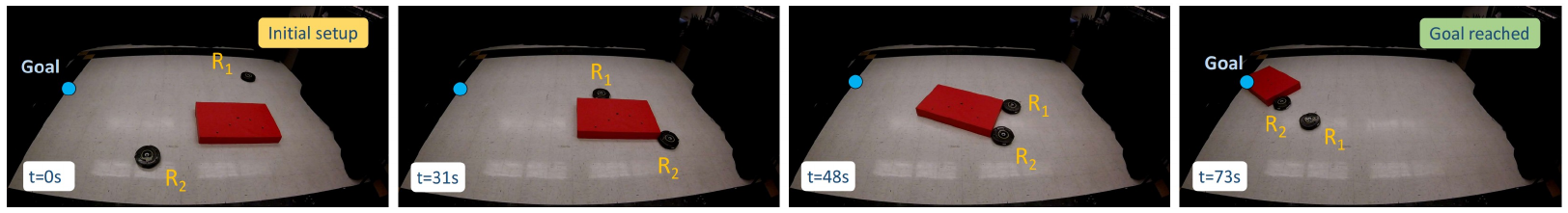}
            \vspace{-6mm}
        \caption{Trial $1$ time lapse illustration of the $2$ robots transporting a rectangular box of uniform weight distribution towards the goal.}
        \label{fig:exptA2_tl}
        \vspace{2mm}
    \end{subfigure}
    \begin{subfigure}[b]{0.32\textwidth}
    \captionsetup{width=.99\linewidth}%
      \includegraphics[width=2.15in,height=1.0in]{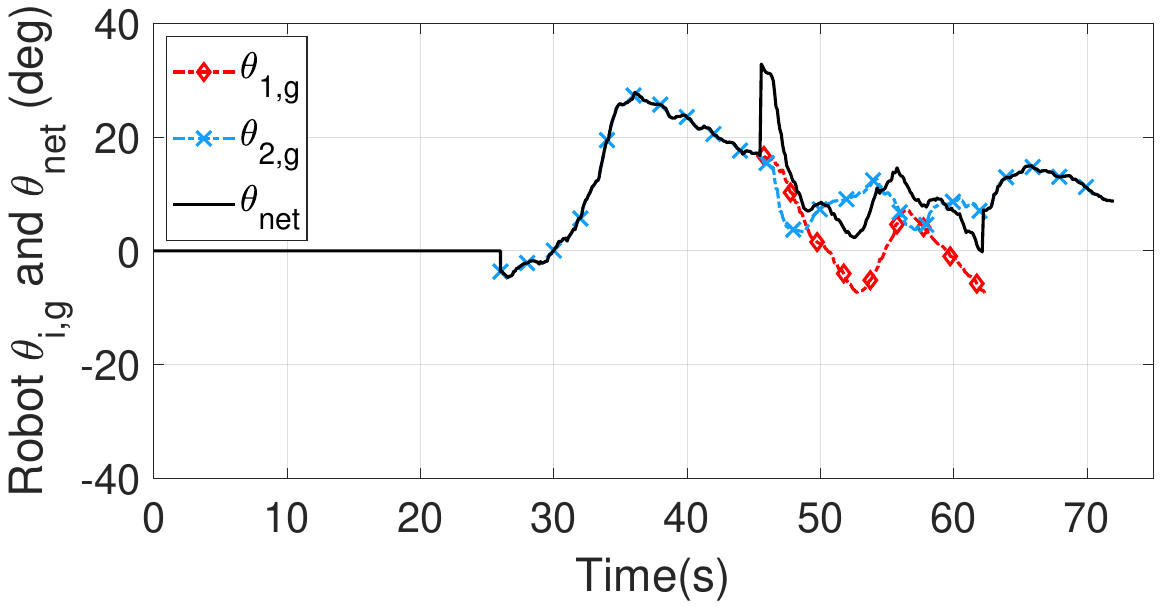}
        \caption{Individual robot and net heading angle error while pushing, confined within $[-\phi, \phi]$ and $[-2\phi, 2\phi]$ respectively.}
        \label{fig:exptA2_theta}
    \end{subfigure}
    \begin{subfigure}[b]{0.32\textwidth}
    \captionsetup{width=.95\linewidth}%
        \includegraphics[width=2.15in,height=1.0in]{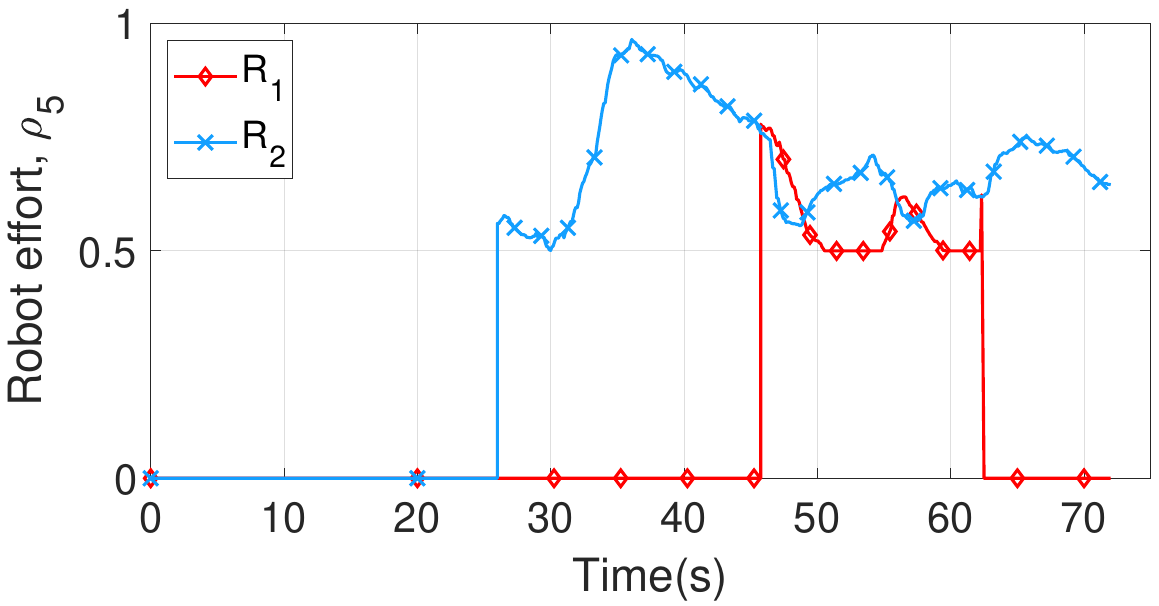}
        \caption{Pushing effort $\rho_5$ in response to individual and collective heading angle error to push object towards the goal location.}
        \label{fig:exptA2_effort}
    \end{subfigure}
    \begin{subfigure}[b]{0.32\textwidth}
    \captionsetup{width=.98\linewidth}%
        \includegraphics[width=2.15in,height=1.0in]{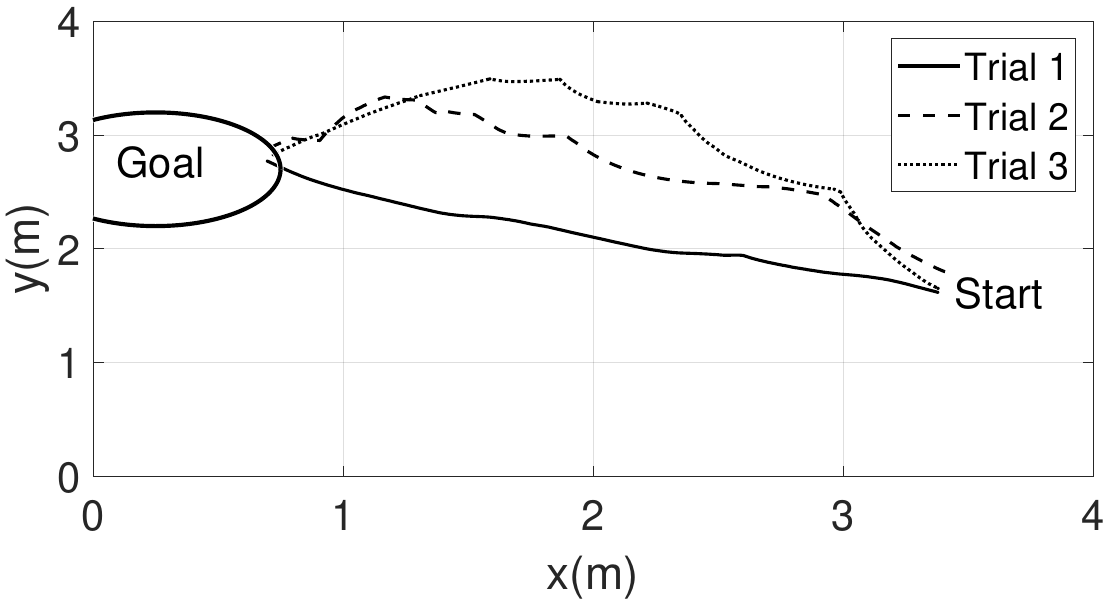}
        \caption{Object centroid path to goal, showing object path converging towards the goal region successfully for all trials.}
        \label{fig:exptA2_path}
    \end{subfigure}
    \caption{Experiment A.2: Transporting a $1.27\times 0.78$\,m, $3.5$\,kg rectangular box with uniform weight distribution using $2$ robots. Due to its shape, two pushing locations are available when a side is perpendicular to the goal direction; at other times, the box must be pushed at the corners. The robots are observed to collaboratively transport the object to the goal.}
    \vspace{-4mm}
    \label{fig:exptA2}
\end{figure*}
The proposed multi-robot object transportation method is validated over three sets of experiments. 
Sets A and B present object transportation results for a disk and a rectangular box demonstrating the effectiveness of our object-model-free pushing effort regulation maintaining robustness to dynamic changes in weight and center of mass. 
Set C validates the system's effectiveness in transporting complex, non-convex objects with non-uniform weight distribution. 
In all trials, all robots were assumed capable of pushing the object individually, had no prior knowledge of the object's properties or the number of robots involved. 
Performance was measured as elapsed transportation time and path efficiency defined as $\frac{l_{min}}{l}$; $l_{min}$ is the shortest distance between object centroid at initial time and the nearest point on a defined goal region, and $l$ is the actual distance travelled by the object centroid. 
A supplementary video is provided with all real-world experiments and simulations \url{https://youtu.be/C2usYt0xzl8}.

\subsection{Experiment A: Disk and Rectangular Box Transportation}\label{sec:caseA}
In Experiment Set A, two iRobot Create platforms use their front bumpers to transport a uniform $0.04$\,kg disk of diameter $0.83$\,m (Expt. A.1) and a $3.5$\,kg rectangular box sized $1.27 \times 0.78$\,m (Expt. A.2) across a $4 \times 4$\,m level surface. 
A VICON tracking system handles positioning and local object boundary detection within limited FOV. Geometrically, the disk offers a single potential pushing location, whereas the box presents two at specific orientations.

Figures \ref{fig:exptA1_tl} and \ref{fig:exptA2_tl} illustrate the transportation process from initial random deployment $t=0$\,s to a goal approximately $3.30$\,m away from the payload centroid. Successful transportation requires the object centroid to reach within $0.5$\,m goal radius. 
Execution parameters are configured as: $\rho_1=1$, $\rho_2=\rho_3=\rho_6=0.56$, $\rho_4=0.6$, $\rho_r=0.003$, $v_m=0.3$\,m/s, $\omega_m=2.2$\,rad/s, $d_r=2$\,m, $d_e=0.3$\,m, $d_s=0.3$\,m, $\alpha=1$, $\delta=0.2$ and $\phi=30^o$.

In Experiment A.1, the robots are observed to follow the boundary at $t=17$\,s. Having successfully detected the potentially single pushing location, the disk was transported by a single robot $R_2$ as observed in $t=34$\,s and $t=62$\,s. With no prior knowledge of the object shape, $R_1$ continued to follow the boundary looking for a pushing location. Fig. \ref{fig:exptA1_theta} and \ref{fig:exptA1_effort} plots the measured individual heading angle error of each robot and the net heading error governing $\eta$ determining the boundary following direction, and the resulting pushing effort. Without $R_1$ ever pushing the object, $\theta_{net}$ was equal to $\theta_{2,g}$, and $R_2$ is observed to regulate its pushing based on the heading angle error. Since $R_1$ was never involved in pushing, its pushing effort remained zero throughout. 
The disk transportation process was repeated independently for two additional trials. The object centroid path from the start to the goal location for the $3$ trials are shown in Fig. \ref{fig:exptA1_path}; the disk was successfully transported in all trials with the lowest path efficiency recorded to be $96$\,\%; raw transport times were recorded as $\{62.1, 60.3, 73.8\}$\,seconds.

Similar observations were made for the rectangular box transportation in Experiment A.2. The measured individual heading angle error of each robot and the net heading error along with the resulting pushing effort are plotted in Fig. \ref{fig:exptA2_theta} and \ref{fig:exptA2_effort}. $R_2$ initially identified the corner of the box as a potential pushing location at $t=26$\,s. It continued to push regulating its pushing effort based on $\theta_{net}$ equal to $\theta_{2,g}$ until the resultant change in orientation of the box allowed a second pushing location of the object to become available to $R_1$ at $t=47$\,s. The two robots were then observed to successfully transport the rectangular box collaboratively to the set goal location. $R_2$ was continuously observed to exert a larger pushing effort to generate a net clockwise rotation of the object around its unknown center of mass as $\theta_{net}$ remained positive throughout due to the goal location being on the opposite end of the center of mass of the box from $R_2$. $R_1$ was observed to correct any overshoots in the clockwise rotation of the object by balancing it with an increasing counterclockwise motion when its own heading angle error became the same sign as $\theta_{net}$. As a result, a fairly straight efficient path of the object was observed to the goal location in Trial 1 as shown in Fig. \ref{fig:exptA2_path}. 
\begin{figure*}[t]
    \centering
    \begin{subfigure}[b]{0.32\textwidth}
    \captionsetup{width=.95\linewidth}%
      \includegraphics[width=2.15in,height=1.0in]{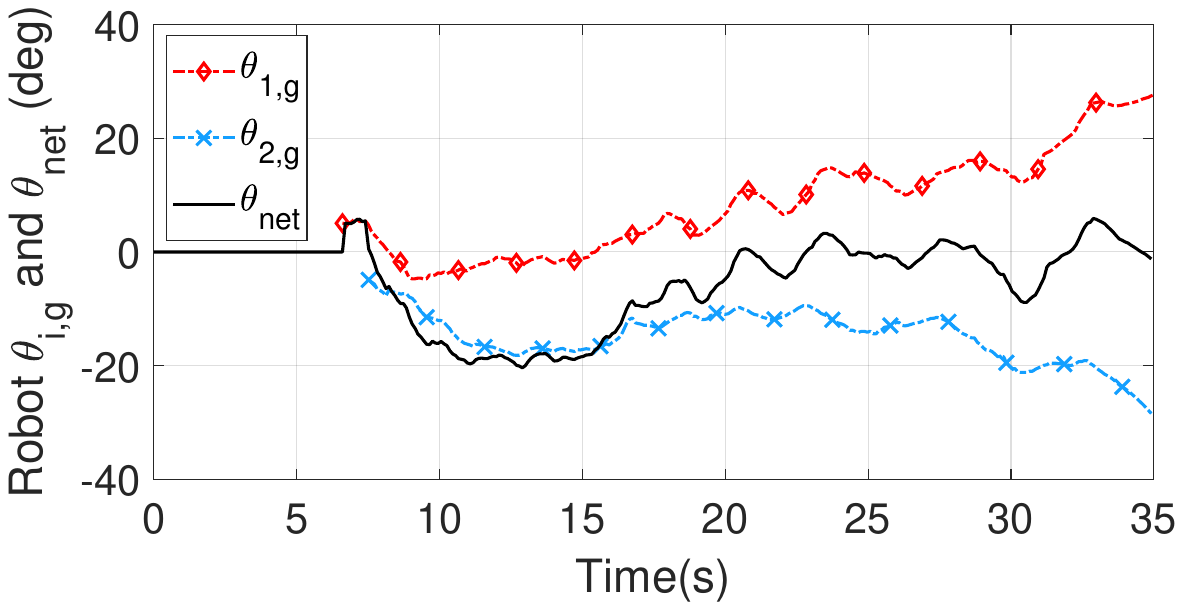}
        \caption{Individual robot and net heading angle error while pushing, confined within $[-\phi, \phi]$ and $[-2\phi, 2\phi]$ respectively.}
        \label{fig:exptB1_theta}
    \end{subfigure}
    \begin{subfigure}[b]{0.32\textwidth}
    \captionsetup{width=.95\linewidth}%
        \includegraphics[width=2.15in,height=1.0in]{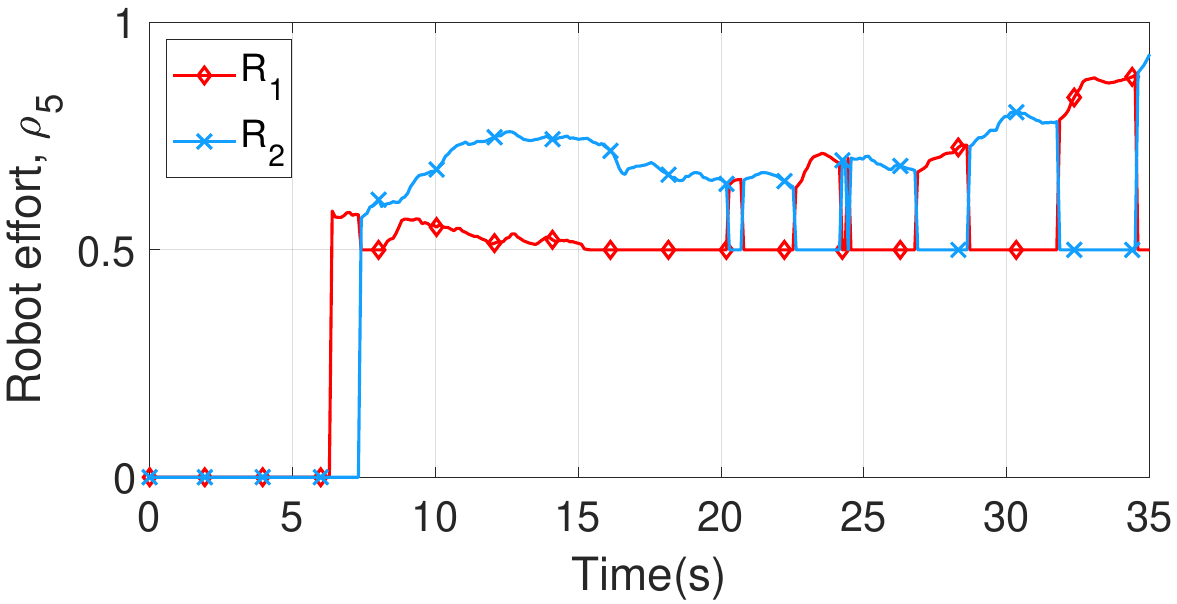}
        \caption{Pushing effort $\rho_5$ in response to individual and collective heading angle error to push object towards the goal location.}
        \label{fig:exptB1_effort}
    \end{subfigure}
    \begin{subfigure}[b]{0.32\textwidth}
    \captionsetup{width=.98\linewidth}%
        \includegraphics[width=2.15in,height=1.0in]{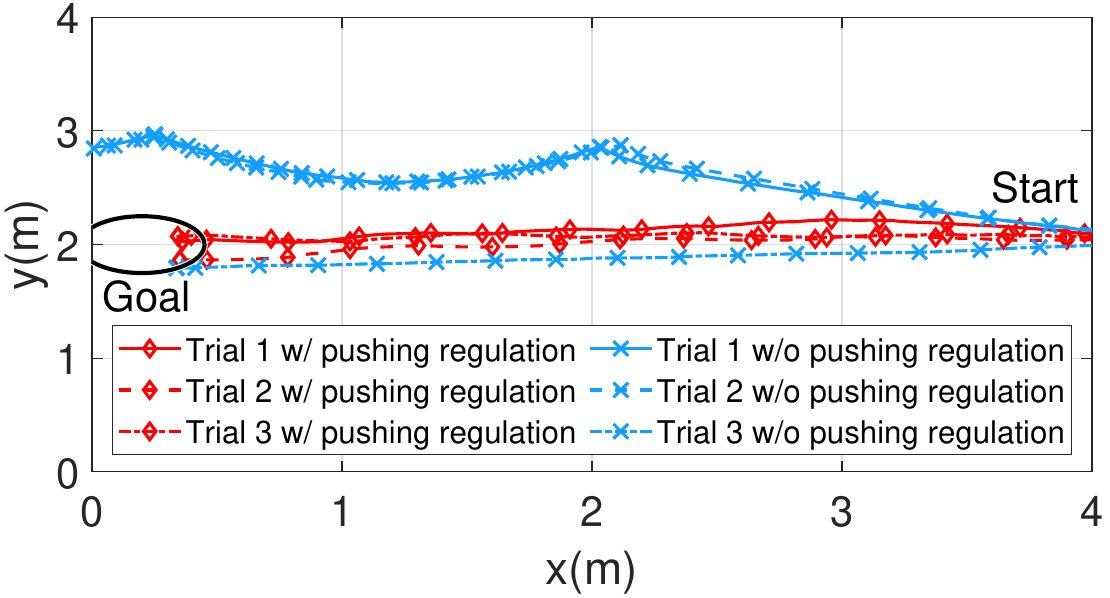}
        \caption{Object centroid trajectories with and without pushing effort regulation, showing larger path deviations without regulation.}
        \label{fig:exptB1_path}
    \end{subfigure}
    \caption{Experiment B.1: Two robots placed to find two pushing locations simultaneously; the rectangular box is initially positioned such that the object center to goal vector is perpendicular to its longest edge. The proposed pushing effort regulation resulted in efficient straight paths to the goal over all trials compared to a constant pushing effort.}
    \vspace{-4mm}
    \label{fig:exptB1}
\end{figure*}
\begin{figure*}[t]
\centering
    \begin{subfigure}[b]{0.32\textwidth}
    \captionsetup{width=.95\linewidth}%
      \includegraphics[width=2.15in,height=1.0in]{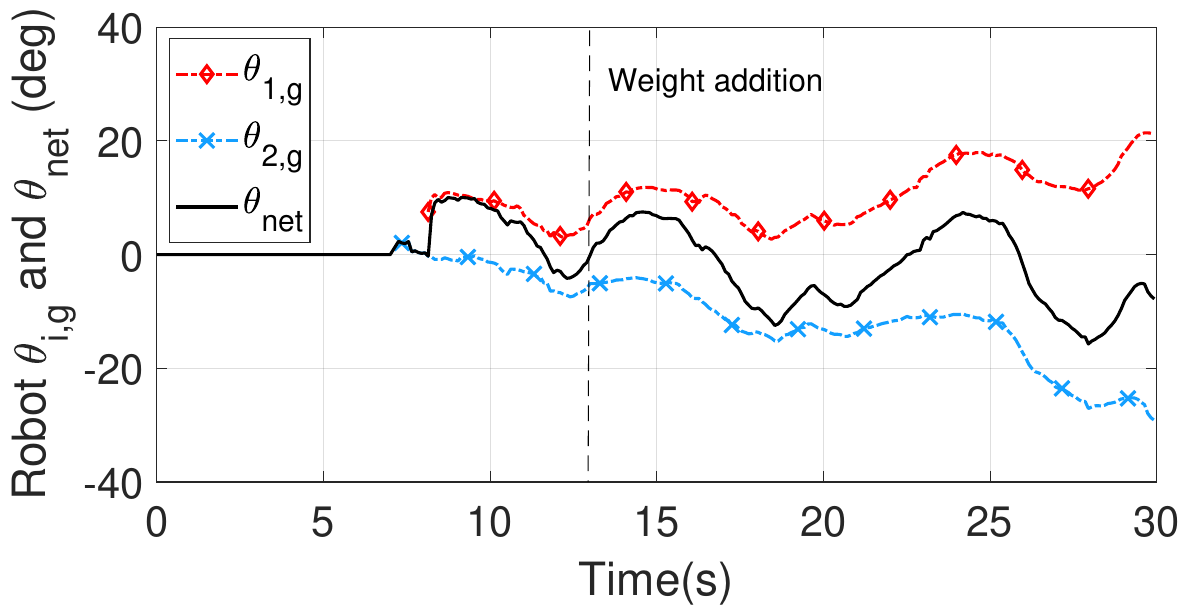}
        \caption{Individual robot and net heading angle error while pushing, confined within $[-\phi, \phi]$ and $[-2\phi, 2\phi]$ respectively.}
        \label{fig:exptB2_theta}
    \end{subfigure}
    \begin{subfigure}[b]{0.32\textwidth}
    \captionsetup{width=.95\linewidth}%
        \includegraphics[width=2.15in,height=1.0in]{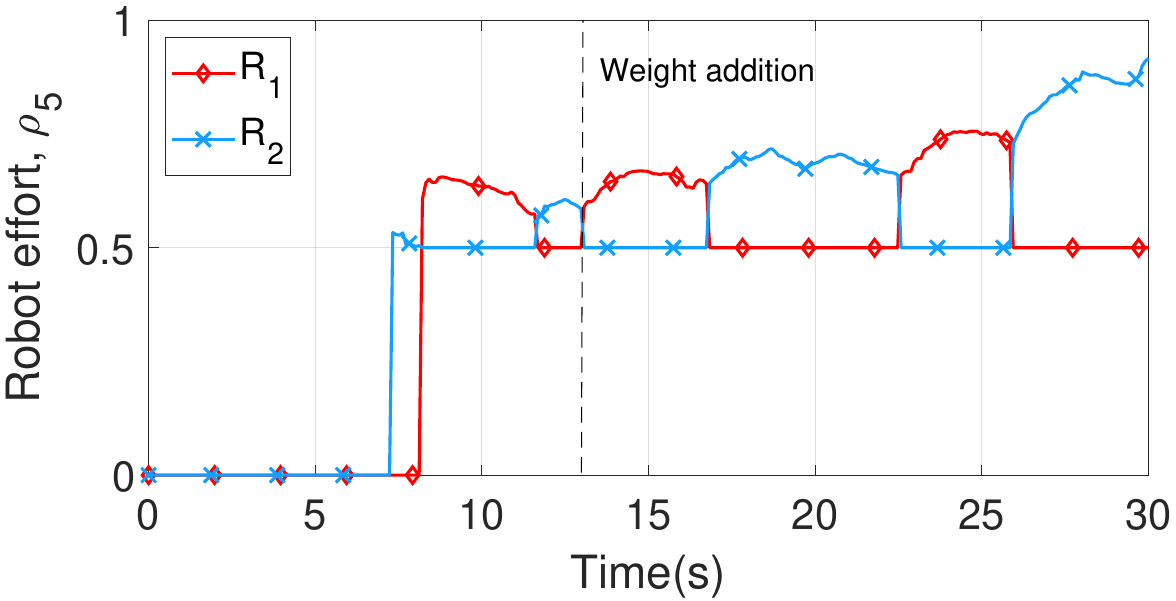}
        \caption{Pushing effort $\rho_5$ in response to individual and collective heading angle error to push object towards the goal location.}
        \label{fig:exptB2_effort}
    \end{subfigure}
    \begin{subfigure}[b]{0.32\textwidth}
    \captionsetup{width=.98\linewidth}%
        \includegraphics[width=2.15in,height=1.0in]{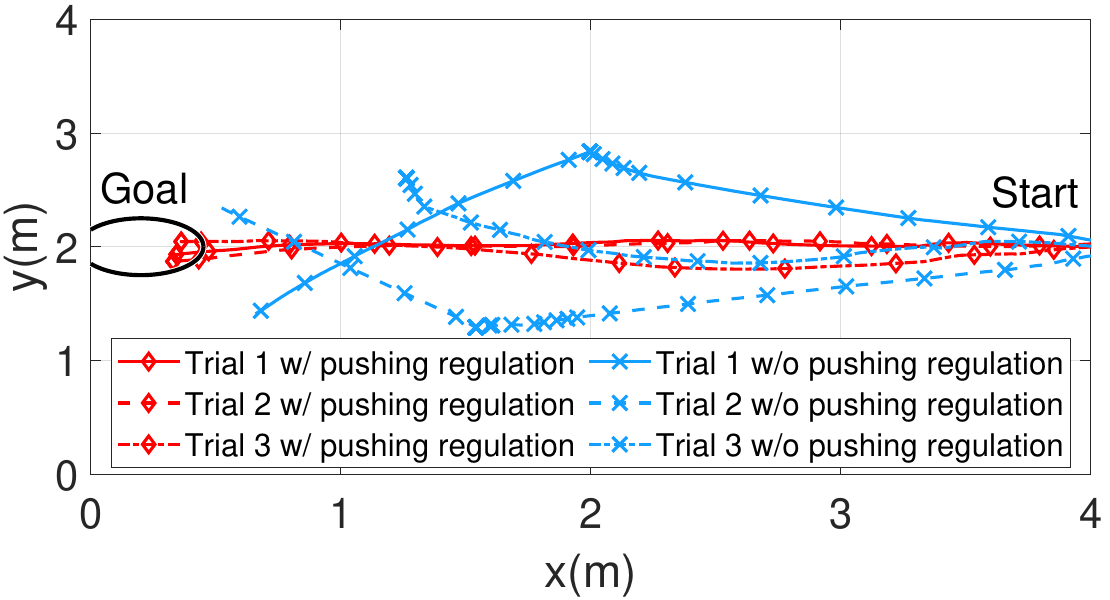}
        \caption{Object centroid trajectories with and without pushing effort regulation, showing larger path deviations without regulation.}
        \label{fig:exptB2_path}
    \end{subfigure}
    \caption{Experiment B.2: An additional weight of $2.2$\,kg added on one end of the object during the straight line transportation process as in Experiment B.1, to mimic dynamically changing weight and weight distribution of the object. The proposed pushing effort regulation resulted in efficient straight paths to the goal over all trials compared to a constant pushing effort.}
    \vspace{-5mm}
    \label{fig:exptB2}
\end{figure*}
The rectangular box transportation process was repeated independently for two additional trials. Although the object was successfully transported to the goal location in all trials, some of the observed paths resulted in a large standard deviation of required transport time, due to the shape of the rectangular box. For a majority of the time, the robots could push the object from corner locations only; for more than one pushing location to be used, one of the boundary surfaces of the object needed to be perpendicular to the goal vector from a robot currently following this side of the object boundary; which only happened in rare instances with the $2$ robot setup. The process could therefore be improved with redundant robots available. Regardless, the lowest path efficiency was observed to be $71.2\%$. Paths recorded from trials $2$ and $3$ are shown in Fig. \ref{fig:exptA2_path}. 
The rectangular box shaped object raw transport times were recorded as $\{73.2, 85.9, 129.7\}$\,seconds. 

\subsection{Experiment B: Robustness of Object-Model-Free Pushing Effort Regulation}\label{sec:caseB}
Experiment Set B setup includes two iRobot Create robots and the rectangular box from Experiment A.2. 
The rectangular box is placed such that its long edge is perpendicular to the goal vector, with robots positioned on opposite sides of the rotational pivot.
Experiment B.1 demonstrates balanced pushing effort regulation to maintain a straight line path towards the goal robust to any wheel slip and irregular friction effects; Experiment B.2 validates the robustness of pushing effort regulation to dynamic weight changes by adding a $2.2$\,kg weight on one end of the box during transport, dynamically shifting its center of mass.

In Experiment B.1, both robots identified pushing locations within $10$\,s, positioning themselves on opposite sides of the object's rotational pivot. 
Consequently, heading errors $\theta_{1,g}$ remained positive and $\theta_{2,g}$ remained negative keeping $\theta_{net}$ bounded within $[\theta_{2,g},\theta_{1,g}]$. 
The robots regulated $\rho_5$ individually to drive $\theta_{net}$ toward zero, countering any drift from wheel slip or friction effects. 
Because $\rho_5$ always remained within the maximum pushing capability of the robots with $\theta_{i,g}$ for $i \in \{1,2\}$ remaining within $\pm \phi$, no robot relocations were required. 
The heading angle errors and the pushing effort are shown in Fig. \ref{fig:exptB1_theta} and \ref{fig:exptB1_effort} respectively.
The path traced by the geometric centroid of the object during transportation shows a fairly straight line with minor deviations over $3$ presented trials as shown in Fig \ref{fig:exptB1_path}. 
The cross-track error of the object centroid path with the shortest distance path remained within $0.12$\,m at all times with the lowest path efficiency recorded as $93.5$\,\%. 
For comparison, Experiment B.1 was repeated without the proposed pushing effort regulation; robots pushed with a constant pushing effort of $0.5$ within a heading angle error of $[-\phi, \phi]$. 
The resulting object paths from $3$ trials are shown in Fig. \ref{fig:exptB1_path}.
Trial $2$ achieved a relatively straight path with a maximum cross-track error of $0.22$\,m; trials $1$ and $3$ failed due to initial contact timing differences and irregular friction, causing the object to drift off course.
Without rotation correction from pushing effort regulation, the robots pushed until heading errors exceeded $[-\phi, \phi]$ triggering relocations.
Due to experiment setup space limitations, trials $1$ and $3$ were stopped without reaching the goal when the object centroid $x$-position turned negative. 

Similar observations were made in Experiment B.2, where an additional weight of $2.2$\,kg was added on one end of the rectangular box at $t=13$\,s during the transportation process. 
The measured individual heading angle error of each robot and the net heading error along with the resulting pushing effort are shown in Fig. \ref{fig:exptB2_theta} and \ref{fig:exptB2_effort} respectively.
With the dynamic weight addition, the object started to rotate clockwise between $14 \leq t \leq 17$\,s with $\theta_{net}$ turning negative between time $16 \leq t \leq 23$\,seconds.
As $\theta_{2,g}$ decreased with $\theta_{2,g}<0$, $R_2$ increased its pushing effort to $\rho_5=0.7$ to counter, while $R_1$ remained at the minimum $\rho_5=0.5$ with $\theta_{1,g}>0$.
This differential effort successfully rotated the object back forcing  $\theta_{net}$ toward zero.
As the object neared the goal, both robots increased their efforts due to growing individual heading errors, with $R_2$ exerting a larger pushing effort to compensate for the shifted center of mass. 
Across three trials, the object maintained a fairly straight path as shown in Fig \ref{fig:exptB2_path} with a maximum cross-track error of $0.3$\,m and the lowest recorded path efficiency of $87.7$\,\%.

Without pushing effort regulation, the object exhibited larger path deviations, as shown in Fig. \ref{fig:exptB2_path}. In trials $1$ and $3$, the center of mass offset caused a net clockwise rotation and rightward drift of the object, requiring robot relocation. In trial $2$, both robots settled on the same side of the object's center line to the goal, causing leftward drift. All three trials without pushing effort regulation were stopped before goal attainment, with the object within $1$\,m of the goal region but misaligned and requiring robot relocation. In trial 2, both robots settled on the same side of the object’s center line to the goal causing a leftward drift.
In all three trials, the paths without pushing effort regulation were significantly less efficient; the transportation process was stopped being deemed irrelevant when the object was within $1$\,m of the goal region with inaccurate heading, requiring robot relocation.  

\subsection{Experiment C: Robustness to Arbitrary Object Shape, Weight Distribution and Initial Pose}\label{sec:caseC}
\begin{figure*}[t]
    \centering
    \begin{subfigure}{\textwidth}
      \includegraphics[width=\textwidth, height=0.85in]{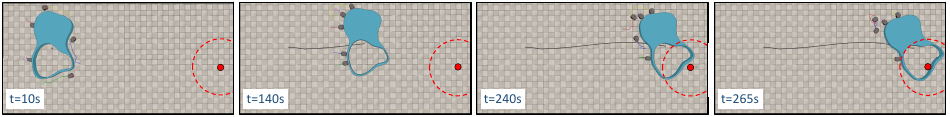}
        \caption{Time lapse illustration of Trial $1$ of the $5$ Pioneer robot arbitrary shaped object transportation process towards the goal (red).}
        \label{fig:exptC_tl}
    \end{subfigure}
    ~
    \begin{subfigure}[b]{0.32\textwidth}
        \includegraphics[width=2.25in,height=1.185in]{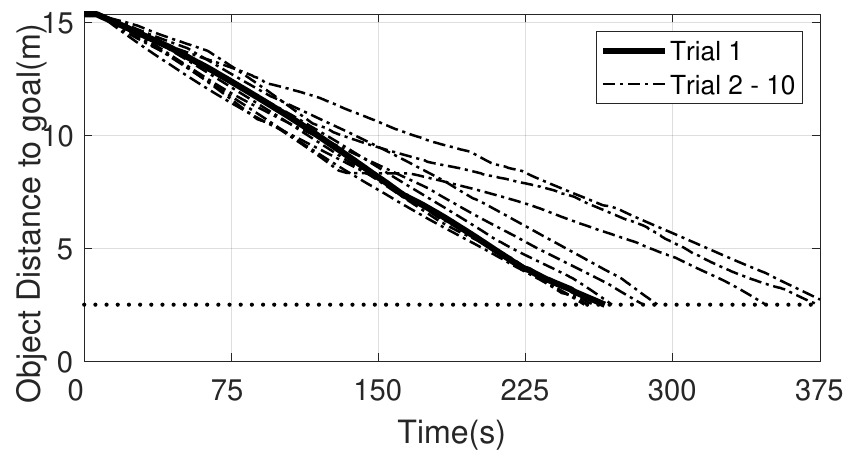}
        \caption{Object centroid distance to goal with time, showing object moves to goal location in finite time for all trials.}
        \label{fig:exptC_objDist}
    \end{subfigure}
    ~
    \begin{subfigure}[b]{0.32\textwidth}
       \includegraphics[width=2.25in,height=1.2in]{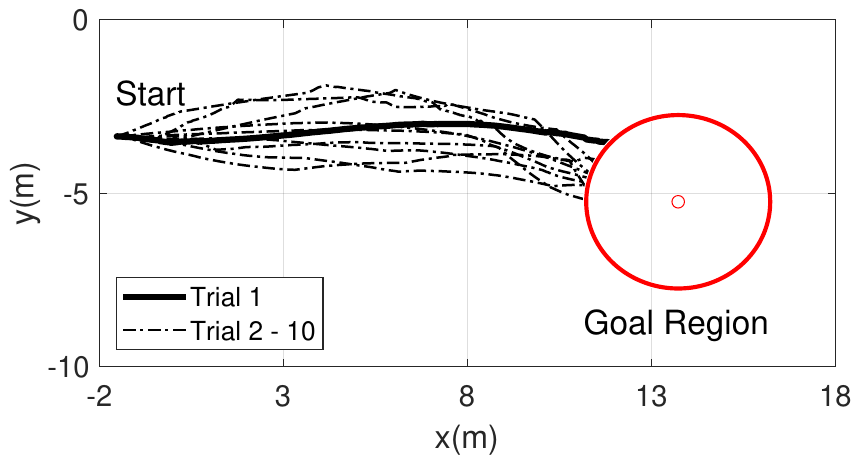}
        \caption{Object centroid path to goal, showing object path converging towards the goal region successfully for all trials.}
        \label{fig:exptC_objpath}
    \end{subfigure}
    ~
    \begin{subfigure}[b]{0.32\textwidth}
        \includegraphics[width=2.15in,height=1.2in]{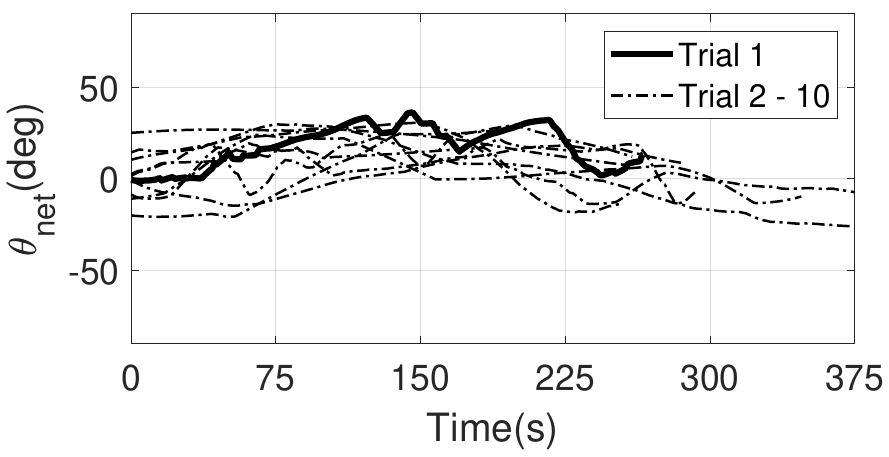}
        \caption{Collective heading error $\theta_{net}$ for all trials remained within their specific bounds throughout the transportation processes.}
        \label{fig:exptC_theta}
    \end{subfigure}
    \caption{Experiment C: Arbitrary shaped object with non-uniform weight distribution at $\theta_o=30^o$ transported to goal location by $5$ robots regulating their pushing effort and relocating if necessary in response to object behavior.}
    \vspace{-6mm}
    \label{fig:exptC}
\end{figure*}
\begin{figure}[t]
\centering
    \begin{subfigure}[t]{0.49\linewidth}
    \centering
    \captionsetup{width=.96\linewidth}
        \includegraphics[width=1.6in,height=1.2in]{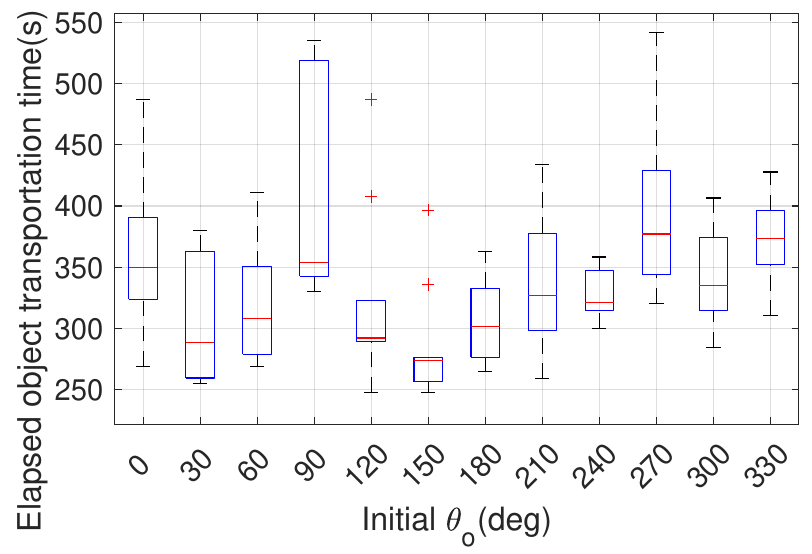}
        \caption{Median and variability of object transportation elapsed time of $10$ trials for each $\theta_o$.}
        \label{fig:robustness_orientation}
    \end{subfigure}
    \begin{subfigure}[t]{0.49\linewidth}
    \centering
    \captionsetup{width=.96\linewidth}
        \includegraphics[width=1.6in,height=1.18in]{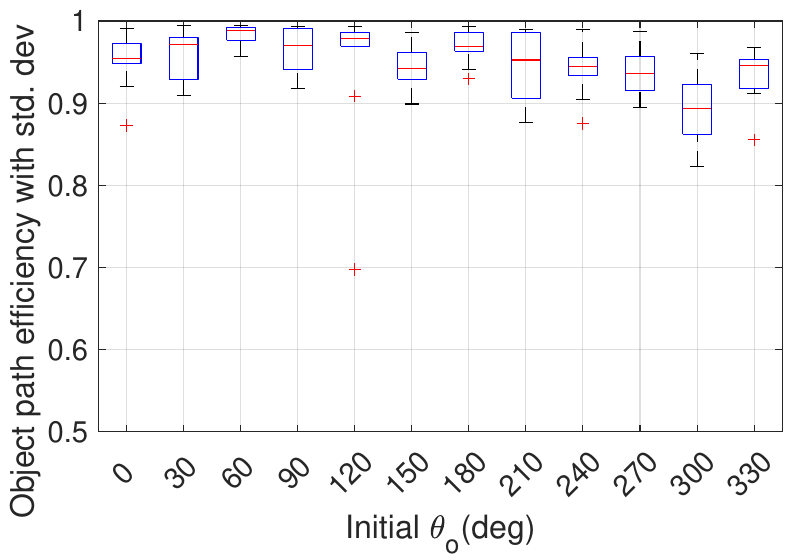}
        \caption{Median and variability of path efficiency of $10$ trials for each $\theta_o$.}
        \label{fig:robustness_pathEfficiency}
    \end{subfigure}
\caption{Object of non-uniform weight successfully transported to defined goal location with $10$ randomly placed robots around the object, for all trials of initial object orientation $\theta_o=0^o$ to $330^o$ at increments of $30^o$.}
\vspace{-7mm}
\label{fig:robustness}
\end{figure}

Experiment Set C is set up with five Pioneer P3-DX robots on a $20 \times 15$\,m planar area on the physics-based 3D robotic simulator V-Rep (CoppeliaSim). 
The robots transported a $10$\,kg arbitrary-shaped object with non-convex boundary regions and its center of mass at an offset from its geometric centroid; the front bumpers were utilized for pushing and the ultrasonic sensors for distance measurements. 
The goal was placed $13.42$\,m away closer to the object’s lighter end, creating a setup prone to counter-clockwise drift. 
Fig. \ref{fig:exptC_tl} shows the time lapse illustration of Trial 1 (object initial orientation $\theta_o=30^o$) of the object transportation process. 
The object path was initially observed to rotate around the heavier end; robots dynamically relocated to the heavier side and regulated pushing effort to correct this drift; the goal was reached in around $4.5$ minutes following a fairly straight line path with minor deviations closer to the goal.
$10$ independent trials were conducted and the resulting object centroid distance to goal, path to goal and the net heading angle error $\theta_{net}$ are shown in Fig. \ref{fig:exptC_objDist}, \ref{fig:exptC_objpath} and \ref{fig:exptC_theta} respectively. 

\begin{figure*}[t!]
\centering
	\includegraphics[width=0.95\linewidth]{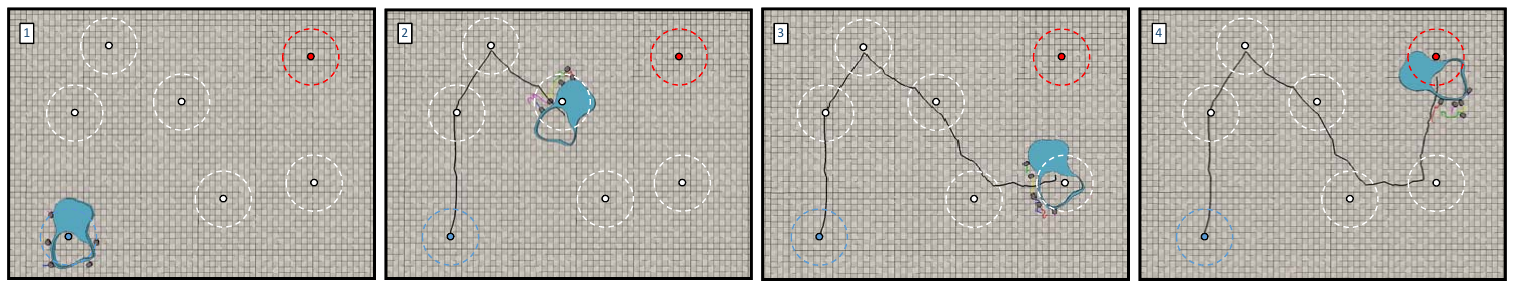}
          \vspace{-2mm}
  	\caption{Object transportation to goal locations extended to waypoint following with acceptable cross-track error. Blue, red and white markers denote starting, end and intermediate path positions.}
    \vspace{-6mm}
\label{fig:path_follow}
\end{figure*}

Experiment sets A and B concluded that the number of available pushing surfaces, their utilization depending on initial robot deployment, and minor differences in the initial object orientation can have a significant impact on the object transportation performance. 
Experiment C was therefore repeated for initial object orientation $\theta_o=$ $0^o$ to $330^o$ at $30^o$ increments with $10$ independent trials at each orientation, and random initial robot placements. The object was successfully transported to the goal location for all trials; box-and-whisker plots of the results are shown in Fig. \ref{fig:robustness_orientation} and \ref{fig:robustness_pathEfficiency}.
The median elapsed time was recorded between $250$ and $400$\,s with the minimum and maximum of $273$ and $378$\,s at $\theta_o=150^o$ and $\theta_o=270^o$ respectively.
The greatest variability in elapsed time occurred at $\theta_o=90^o$ and $270^o$ where the robots were either pushing only the heavier end or the lighter end of the object with fewer available pushing points.
Path efficiency across orientations remained consistently high with medians exceeding $90$\,\% except for one instance of $87$\,\% at $\theta_{o}=300^o$.
The results validate that the proposed system effectively manages complex geometries and non-uniform mass distributions regardless of initial pose.
The proposed system can be used for waypoint following of an object to move around obstacles; a successful implementation by $5$ robots is shown in Fig. \ref{fig:path_follow}. 

\begin{figure}[t]
\centering
    \begin{subfigure}[t]{0.485\linewidth}
    \centering
    \captionsetup{width=.96\linewidth}
        \includegraphics[width=1.6in,height=1.15in]{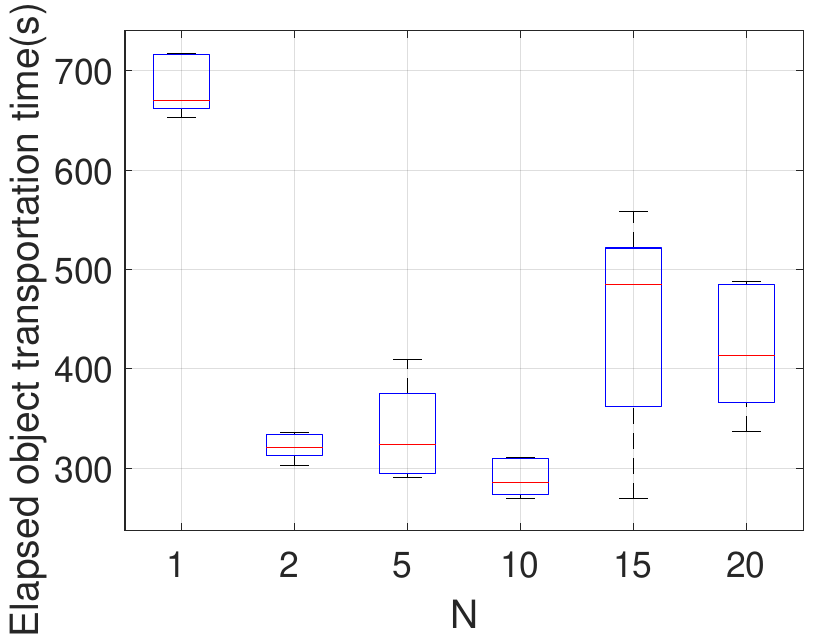}
        \caption{Median and variability of object transportation elapsed time of $10$ trials for each $N$.}
        \label{fig:scalability_objDist_mean}
    \end{subfigure}
    \begin{subfigure}[t]{0.485\linewidth}
    \centering
    \captionsetup{width=.96\linewidth}
        \includegraphics[width=1.6in,height=1.15in]{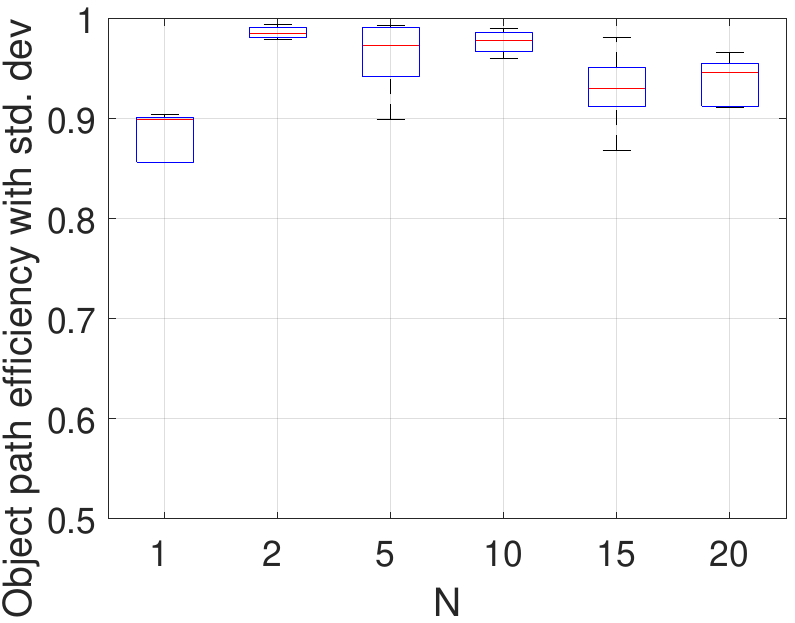}
        \caption{The median and variability of path efficiency for $10$ trials for each $N$.}
        \label{fig:scalability_pathEfficiency}
    \end{subfigure}
\caption{Scalability: The object of non-uniform weight with $\theta_o=30^o$ was successfully transported to the defined goal with $N=1$, $2$, $5$, $10$, $15$ and $20$ randomly placed robots around the object over $10$ trials each.}
 \vspace{-7mm}
\label{fig:scalability}
\end{figure}

\subsection{Scalability Analysis}\label{sec:scalability}
While larger robot groups can potentially provide greater net pushing effort, object transport performance is limited by available pushing surfaces and maintaining inter-robot safety distances. 
Experiment C with the arbitrary shaped object initially at $\theta_o = 30^o$ was repeated using $N=1$, $2$, $5$, $10$, $15$, and $20$ Pioneer P3-DX robots for $10$ independent trials each with random initial robot placements. 
The median and variability of the object transportation elapsed time and path efficiency are presented in Fig. \ref{fig:scalability_objDist_mean} and \ref{fig:scalability_pathEfficiency} respectively. $N=1$ yielded the longest transportation time and decreased as the group grew to $N=10$. 
However, for $N>10$ transport times increased due to constant robot relocations to maintain inter-robot minimum safety distance as their pushing surfaces slide along the object boundary during its motion.
This was confirmed with the scalability experiment repeated for the same object $3$ times the current size; the lowest elapsed time was obtained for $N=20$. 
Across all cases, median path efficiency was above $90$\,\% with the lowest observed for $N=1$.
These findings confirm that optimal team size is intrinsically linked to the physical dimensions of the object.

\section{CONCLUSION}\label{sec:conclude}
\vspace{-1mm}
This paper presented a decentralized, model-free multi-robot pushing framework combining a finite-state machine with real-time effort regulation. 
Physical and simulated experiments validated robust, goal-directed transport of arbitrary geometries under non-uniform surface friction, wheel slip, and sudden center-of-mass shifts. 
Transport efficiency depends on the available pushing surfaces and the chosen inter-robot safety distance. 
Future work will integrate anti-slip awareness directly into the effort model, extend the framework to heterogeneous teams, adapt the control laws to 3D manipulation tasks, and relax assumptions regarding rigid-body dynamics and continuous heading estimations.

\addtolength{\textheight}{-5cm}   


\bibliographystyle{IEEEtran}
\bibliography{references}

\end{document}